\documentclass[Journal,letterpaper]{ascelike-new}
\usepackage[utf8]{inputenc}
\usepackage[T1]{fontenc}
\usepackage{lmodern}
\usepackage{graphicx}
\usepackage[figurename=Fig.,labelfont=bf,labelsep=period]{caption}
\usepackage{subcaption}
\usepackage{amsmath}
\usepackage{newtxtext,newtxmath}
\usepackage[colorlinks=true,citecolor=red,linkcolor=black]{hyperref}
\NameTag{Zaker Esteghamati et al, \today}
\begin{document}
\nolinenumbers

\title{Beyond development: Challenges in deploying machine learning models for structural engineering applications}

\author[1]{Mohsen Zaker Esteghamati}
\author[2]{Brennan Bean}
\author[3]{Henry V. Burton}
\author[4]{M.Z. Naser}
\affil[1]{Department of Civil and Environmental Engineering, Utah State University, Logan, UT. Email: mohsen.zaker@usu.edu}
\affil[2]{Department of Mathematics and Statistics, Utah State University, Logan, UT.}
\affil[3]{Department of Civil and Environmental Engineering, University of California at Los Angeles, Los Angeles, CA.}
\affil[4]{School of Civil and Environmental Engineering and Earth Sciences, Clemson University, Clemson, SC.}

\maketitle

\begin{abstract}

Machine learning (ML)-based solutions are rapidly changing the landscape of many fields, including structural engineering. Despite their promising performance, these approaches are usually only demonstrated as proof-of-concept in structural engineering, and are rarely deployed for real-world applications. This paper aims to illustrate the challenges of developing ML models suitable for deployment through two illustrative examples. Among various pitfalls, the presented discussion focuses on model overfitting and underspecification, training data representativeness, variable omission bias, and cross-validation. The results highlight the importance of implementing rigorous model validation techniques through adaptive sampling, careful physics-informed feature selection, and considerations of both model complexity and generalizability.

\end{abstract}
\section{Introduction}
\label{sec:Intro}
The last decade witnessed a surge in machine learning (ML)-based solutions to address various structural engineering problems, ranging from analyzing \cite{qin2023,xu2022,hwang2021,kourehpaz2022} to designing \cite{zhao2023,esteghamati2021,esteghamati2023all,issa2023} and monitoring \cite{mohammadi2023,soleimani2023,yuan2023,bashar2022} the built environment. The current practice of structural engineering-oriented model development is heavily focused on identifying the most ``accurate'' ML model. This emphasis is mainly due to the justification of ML as an alternative to complex and time-consuming methods, where ML provides a more accurate solution at lower computational cost and time. 

\begin{figure}[t]
   \centering
        \includegraphics[width=15 cm]{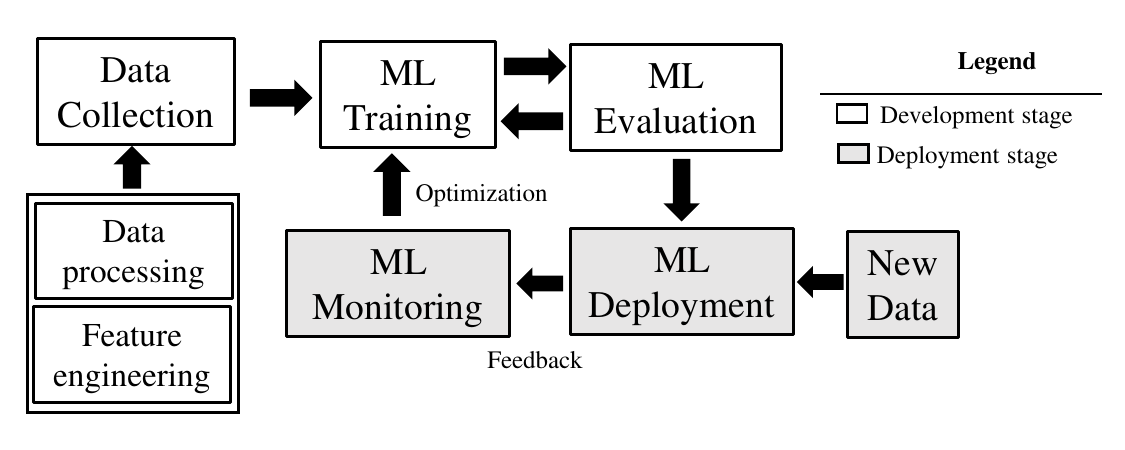}
    \caption {Standard ML pipeline: white boxes show the development stage, whereas grey boxes show the deployment stage. }
    \label{MLpipe}
\end{figure}

The life cycle of an ML model has two main stages of development and deployment (Figure \ref{MLpipe}). Most research papers rightfully focus on model development, though the practical benefits of the ML model are achieved during deployment. However, despite the successful implementations of several ML-based solutions for various structural engineering problems, little attention has been given to exploring the efficacy of these models beyond  "establishing a proof-of-concept". Such explorations are critical as the successful development of ML models does not necessarily translate into useful solutions that can be deployed for real-world datasets \cite{baier2019}. 

 Fundamentally, almost all ML models (due to their statistical nature, and similar to other well-established approaches such as empirical analysis) capture \textit{data association} rather than \textit{causal relationships}. Here, data association refers to one variable providing information about another variable. In contrast, a causal relationship occurs when one variable results from another variable, and its absence leads to a counterfactual statement \cite{naser2023,naser2022,burton2023causal,burton2023}. This means that the variables identified by an ML model as ``important'' for accurate predictions do not necessarily are variables that drive the engineering process in question. Even when ignoring the limitations of ML models in identifying causal relationships, many ML models make it difficult to determine the nature of the association between variables. Such a "black-box" aspect negatively affects user confidence when applying the ML model in deployment. 

 The over-reliance on accuracy metrics during model development might be inadequate, or even problematic, for deployment under specific circumstances. In classification problems on imbalanced datasets (i.e., the proportion of observations of each class type differs substantially), an accurate classifier often discriminates relatively larger class types better, which might not reflect the class types observed in real-world data. For highly imbalanced datasets, such as ones providing structural failure, careful feature selection is needed to ensure model robust performance \cite{koh2022}. Regression-based ML models can also show high values for accuracy metrics (e.g., $R^2$ or error-related metrics), but fail at deployment, though this issue has been less explored in the structural engineering context. Among various explanations for the disconnect between ML development and deployment, a usual culprit is the over-sampling of the most accessible scenario in the training sets, which results in the model's lackluster performance for scenarios beyond the training regime. 

The aforementioned critical issues raise concerns about the reliability of ML-based solutions for deployment, as the structural engineering discipline historically favors explainable models that can be easily generalized for real-world applications. Part of this concern is due to the mechanistic roots of the field. Therefore, this paper reviews themes from various ML engineering fields, and maps them to the challenges in ML development that adversely impact deployment in the structural engineering domain. In particular, we investigate the possible reasons for the challenges of generalizability (extrapolation beyond the training set) and ML explainability through feature importance. Two illustrative examples are presented to demonstrate each challenge in the context of typical simulation and experimental datasets used in structural engineering problems. The presented material and demonstrations aim to inform structural engineers developing ML models of possible deployment pitfalls, paving a path towards more relevant and reliable ML solutions.  

\section{Challenges of developing ML for deployment}
\label{sec:challenge}
ML models that are applied in the real world must behave in a way that can satisfy deployment needs, with behavioral requirements are domain- and context-specific. As shown in Figure \ref{MLpipe}, a standard ML development pipeline includes model specification, training data, and an independent evaluation procedure. However, such a pipeline often disregards the differences between the evaluation procedure and deployment conditions. This disparity is commonly attributed to \textit{structural flaws} in the design of the pipeline \cite{d2020}, such as selection bias, unavailability of predictors at the time of deployment, or even concept drift (i.e., change in the distribution of input data over time). However, even when structural flaws are avoided, accurate ML models can provide inconsistent, inaccurate, or inexplicable predictions when deployed.

\subsection{Challenge 1: Generalizability beyond the training set}
\label{Generalizability}
ML models may need to make inferences on observations with measurements outside the range of those observed in the training set during deployment. This concept, referred to as generalizability, is commonly measured by testing model performance over a new dataset. However, the limitations of ML models for input space beyond the one sampled in the training data are well-understood \cite{luo2023,luo2021}. This issue of generalizability becomes prominent for deployment, where the real-world data could potentially come from a different data distribution than the one observed in the training set.   

\begin{figure}[t]
   \centering
        \includegraphics[width=12 cm]{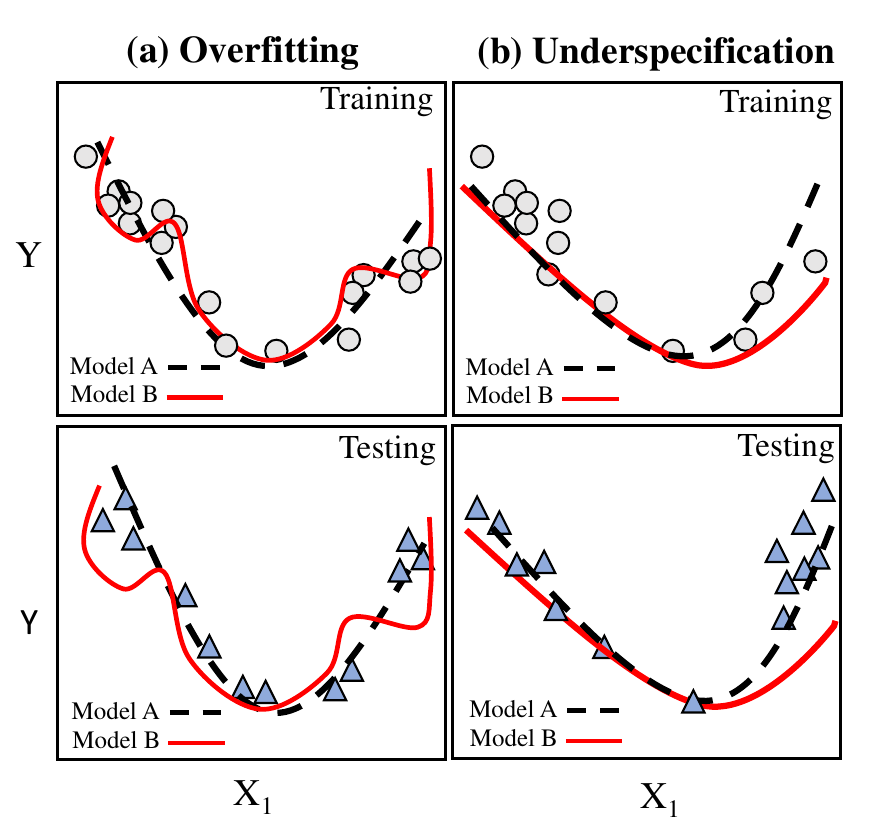}
    \caption {Two common pitfalls of ML models for deployment. (a) overfitting: Model B overfits data on the training set, which results in capturing pseudo relationship (i.e., curves) at lower and higher response values  (b) underspecification: Model B is underspecified as shown by not capturing $X_1$ relationship with the response at high $X_1$ values , hence failing to perform well on the testing set where most of the data occurs at this data region}
    \label{MLpitfall}
\end{figure}

A common issue for the ML model's generalizability is \textit{overfitting}, where the model captures pseudo-relationships or irrelevant patterns in the training data, leading to lower performance on new data (Figure \ref{MLpitfall}.a). Overfitting can occur due to various reasons, such as (i)  small and unrepresentative training data, or (ii) the relationship between model complexity and its predictive power, which position more complex models to have lower bias but higher variance (lower consistency) on new data. Cross-validation and hold-out methods (two-way or three-way) address the former, whereas reducing model complexity (fewer predictors or simpler algorithms) can address the latter.

The inadequacy of training datasets to represent the diversity of the real-world data often leads to poor performance of ``accurate'' ML models in deployment. Besides the size of the dataset, oversampling or undersampling of important scenarios would lead to bias that will propagate into ML deployment. The poor representativeness could lie in the selected features, or the difference between data domains of training and testing sets \cite{mathrani2021,pan2009}. The latter is also referred to as "negative transfer" in the transfer learning literature, where learning from different domains will degrade the performance of the target domain \cite{ge2014}. Lastly, representation bias can occur by compiling accessible data (e.g., average reported values) without consideration of the uncertainties in the compiled value \cite{li2022}. Rigorous evaluation procedures (such as k-fold cross-validation) must be used to tackle these challenges, albeit cross-validation could still lead to biased estimates on smaller datasets \cite{varma2006,vabalas2019}.

\subsection{Challenge 2: Explainability through feature importance}
\label{interpretability}
 Recently, there has been a movement towards explainable artificial intelligence (XAI) in the structural engineering field 
 \cite{esteghamati2023,wakjira2022,naser2022e,cakiroglu2022,feng2021}. Explainability and interpretability are often used interchangeably \cite{allen2023}, although some authors distinguish between their definitions \cite{rudin2022,watson2022}. Efforts have been made to define interpretability in terms of the adopted evaluation method (subjective versus objective) or via other criteria such as sparsity, fidelity of explanation, or sensitivity to perturbations, albeit no universal consensus exists \cite{molnar2020}.  

Among various explainability techniques, feature importance has been a widely practiced method for supervised discoveries \cite{allen2023}. Depending on the selected algorithm, different model-specific feature importance metrics are used, such as loss difference between different splits in tree-based models \cite{allen2023}. Model-agnostic methods, such as feature permutation or Shapely values \cite{shapley1953,lundberg2017}, have also been proposed to provide global explanation irrespective of the selected model. 

While the pursuit of explainable models is certainly desirable, practice suggests that ``not all that glitters is gold.'' Most feature importance metrics aim to demonstrate the consequence of randomly permuting a given variable and measuring the loss of accuracy that results \cite{fisher2019,strobl2008,debeer2020}, rather than measuring the loss of accuracy associated with removing the variable from the model entirely. This means that removing a seemingly important feature (as identified by feature importance) might not actually affect the model's accuracy. In contrast, omitted variable bias (i.e., the absence of an important feature in the model) threatens the underlying causality, and subsequent interpretation of the model \cite{clarke2005,chernozhukov2021}. In addition, feature importance definitions become problematic for correlated features, particularly when permuted-based feature importance methods are used, causing misleading interpretations due to the shared information in several features \cite{hooker2021,strobl2008,molnar2023}.  
 
Another issue affecting explainability (and also generalizability) is "underspecification", which is when chosen features do not \textit{uniquely} explain the desired response ((Figure \ref{MLpitfall}.b). In other words, multiple distinctive sets of features may equally satisfy the evaluation criteria (e.g., accuracy metrics) \cite{d2020}. Several authors define underspecification as the inability of the ML model to capture the causal relationship between response and features, and subsequently, a lack of invariability to confounding factors \cite{eche2021}. This underspecification is a product of flexibility in several choices in the model development pipeline, often directed at automated or randomized search algorithms (e.g., hyperparameter tuning). Furthermore, underspecification cannot be detected when the test set has identical variable distributions as the training set. Current efforts in addressing underspecification involve supplementing the ML pipeline with more stringent evaluation procedures, such as stress tests with shifted evaluation, where the test set is deliberately adjusted to challenge the ML model \cite{eche2021,d2020}. Recently, some authors \cite{yu2022,youssef2023,ramspek2021} argue that no external evaluation could be adequate for deployment, and a continuous local validation (adopted from Machine Learning Operations (MLOps) literature\cite{sculley2015}) is necessary for true generalizability \cite{youssef2023}.        

\section{Illustrative examples}
\label{sec:example}
Two databases were selected from published ML-based studies to use as illustrative examples. The first database contains results of finite element modeling of cold-formed steel (CFS) channels, whereas the second database presents experimental studies on reinforced concrete (RC) walls \cite{aladsani2022}. The selected databases represent two common data types (i.e., simulation \& experimental) used in most ML-based studies in structural engineering. The presented examples represent possible issues that could lead to the discussed deployment challenges. 

The CFS channel database comprises 3512 finite element simulations of ultimate shear strength ($V_{cr}$) of CFS with web slots \cite{degtyarev2021}. The numerical models account for material and geometric nonlinearities and initial imperfections. As such, the database contains information on the following features: channel depth ($D$), flange width ($B$), stiffener length ($B_1$), thickness ($t$), length of slots ($L_{sl}$), the height of slots ($W_{sl}$), spacing of slots in the longitudinal direction ($S_{sl}$), spacing of slots in the transverse direction ($B_{sl}$), number of perforated regions ($N$), number of slot rows ($n$), steel yield stress ($f_y$), type of boundary conditions, inside bend radius ($r$), the aspect ratio ($ a\/h$), and height of the longitudinal stiffener ($h_{st}$).

The RC wall database contains 164 experiments compliant (partially or fully) with provisions of ACI 318-14 for special walls. The database contains information on the following wall parameters: slenderness ratio ($\lambda_b$), shear stress demand ($\nu_{max}/\sqrt{f'_c}$), ratio of transverse reinforcement spacing to longitudinal reinforcement diameter ($s/d_b$), ratio of provided to required area of the boundary transverse reinforcement ($A_{sh,provided}/A_{sh,required}$), axial load ratio ($P/f'_cA_g$), ratio of centerline distance between laterally supported boundary longitudinal bars to the width of the compression zone ($h_x/b$), longitudinal reinforcement ratio of boundary ($\rho_{l,BE}$), web transverse reinforcement ratio ($\rho_{t,w}$), ratio of neutral axis to wall length ($c/l_w$), length of the confined boundary normalized by wall length ($l_{BE}/l_w$), ratio of tested tensile-to-yield strength of the boundary longitudinal reinforcement ($f_u/f_y$) and wall drift. 

\subsection{Example 1: Overfitting and cross-validation}
Preliminary machine learning benchmarks for the CFS channel dataset are computed using the stressor package \cite{stressor} in R 4.3.1 \cite{R2023}, which calls the PyCaret package \cite{PyCaret} via Python 3.8 \cite{python38}. The benchmark process includes a light automated hyper-parameter tuning of 18 machine learning and regression models and subsequent evaluation of those models via 10-fold cross-validation. The results suggest that regression tree approaches (gradient boosting machine and random forests, among others) vastly outperform traditional or penalized regression approaches, with root mean square errors (RMSE) of around 5,000 compared to an RMSE around 21,000 for the linear models (LMs).

The accuracy of the regression tree approaches illustrates their effectiveness in representing a dataset when provided multiple, highly similar, examples. However, what is often of greater interest to practitioners is the ability of a model to make appropriate predictions within the range of the sample space, but for parameter combinations not necessarily observed in the training. Unfortunately, traditional approaches for assessing model accuracy provide poor representations of this kind of predictive ability. 

\begin{figure}
    \centering
    \includegraphics[width = 0.75\textwidth]{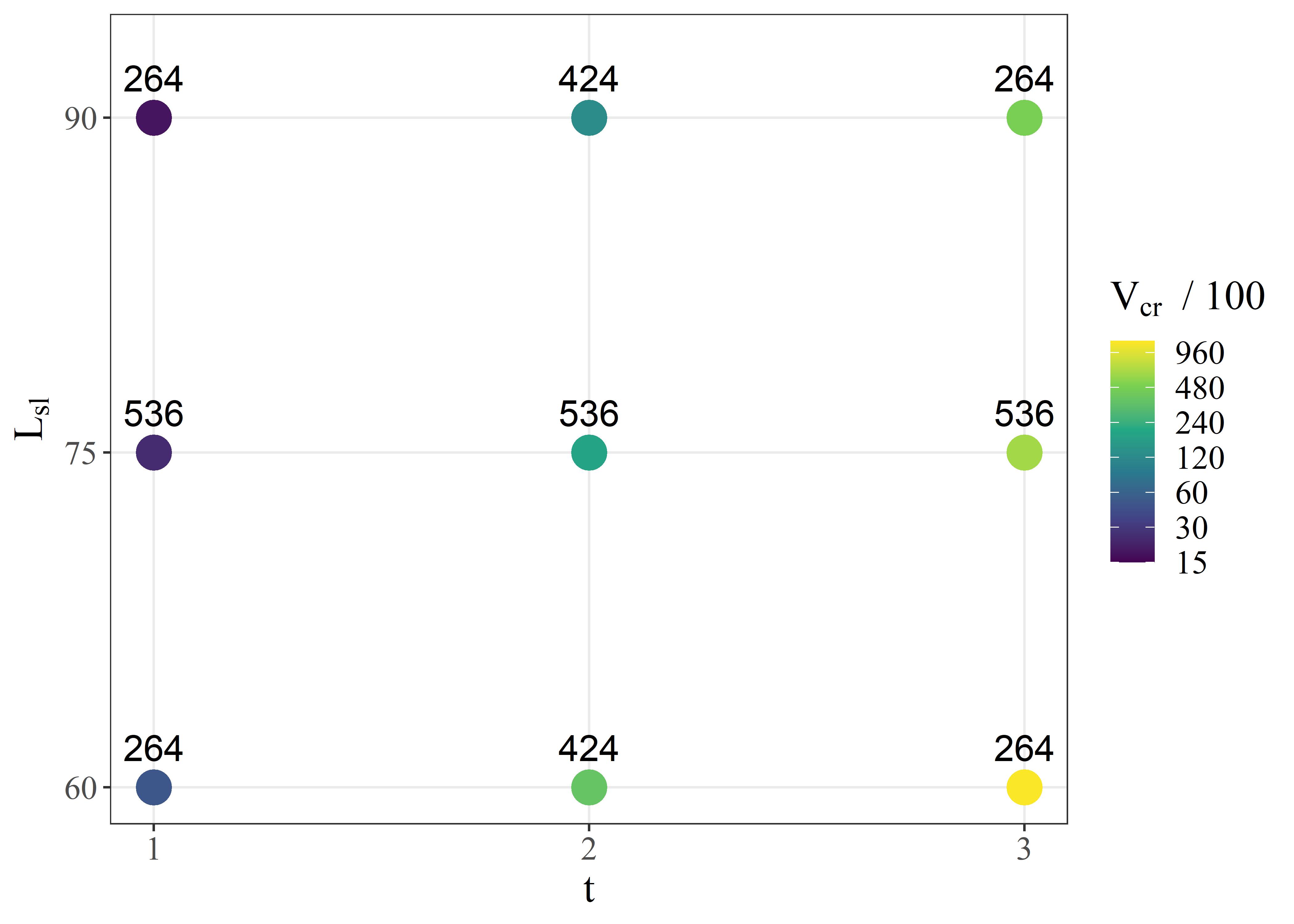}
    \caption{Plot of the unique combinations of $t$ and $L_{sl}$ in the CFS channel dataset, including sample size counts for each group.}
    \label{fig:tlcomb}
\end{figure}

The shortcomings of cross-validation are illustrated though a comparison of the predictive capability of random forest (RF) models to traditional LMs using an adaptation of the cross-validation approach. This adaptation is based on the results of a random forest variable importance analysis, which suggests that $t$ and $L_{sl}$ are far and away the most important variables within the random forest model for predicting $V_{cr}$ . The variables $t$ and $L_{sl}$ are also far and away the most highly correlated with $V_{cr}$. Because these data are simulated, there are exactly nine combinations of $t$ and $L_{sl}$ in the dataset, as shown in Figure \ref{fig:tlcomb}.

\subsubsection{Considered Models}
For both RF models, all available variables are used to predict an un-transformed response variable $V_{cr}$. The difference between the regular and the ``tuned'' model is that the tuned model considers all 14 variables for each partition of the tree (a variable called \texttt{mtry}), while the regular model considers only three.

For the full linear model, all un-transformed explanatory variables are used to predict an un-transformed response with no interaction terms. In summary, the select model considers:
\begin{equation}
    E\left(\log(V_{cr})\right) = \beta_0 + \beta_1t + \beta_2L_{sl} + \beta_{3}(t*L_{sl}). 
\end{equation}

This modified model is a response to the dominating correlations of $t$ and $L_{sl}$ with $V_{cr}$ as well as visuals that suggest a potential two-way interaction between the two variables and $V_{cr}$. These models do not imply that LMs are a appropriate representation of cross sectional capacity. Rather, the LMs simply serve to illustrate the performance of a simple and common modeling approach relative to more complicated ML models.

\subsubsection{Cross-validation} 
Cross-validation provides an alternative to training/test sets for testing model accuracy. The approach randomly separates data into $k$ groups and uses $k - 1$ groups to train the model for testing on the $k^\text{th}$ group. The process is repeated until all groups are used as the test set exactly once. While very popular and reliable for testing model accuracy, the method is prone to providing overly optimistic metrics of model performance on some datasets, such as the one used in this example. 

To illustrate this, cross-validation is performed twice: once using the traditional approach of random selection into groups, and again by assigning group members based upon the observation's combination of $t$ and $L_{sl}$ values. The overall RMSE for two versions of the RF model and two versions of a linear model are provided in Table \ref{tab:rmse}. 

\begin{table}[htbp]
\centering
\caption{Table of summary statistics from traditional and adapted cross-validation for random forest (RF) models and linear models (LMs). Metrics include root mean square error (RMSE) and median absolute error (MAE).}
\begin{tabular}{l|cc|cc}
\hline
Model & \multicolumn{2}{c}{RMSE} & \multicolumn{2}{c}{MAE} \\
 & Traditional & Adapted &  Traditional & Adapted  \\
\hline
RF & 11,927 & 27,240 &  3,763 & 10,055  \\
RF-Tuned & \textbf{5,591} & 37,320 &  \textbf{684} & 14,241  \\
LM-Full & 20,830 & 26,674 &  11,180 & 15,231  \\
LM-Select & 20,356 & \textbf{24,106} &  3,833 & \textbf{4,713}  \\
\hline
\end{tabular}
\label{tab:rmse}
\end{table}

\begin{figure}
    \centering
    \includegraphics[width = \textwidth]{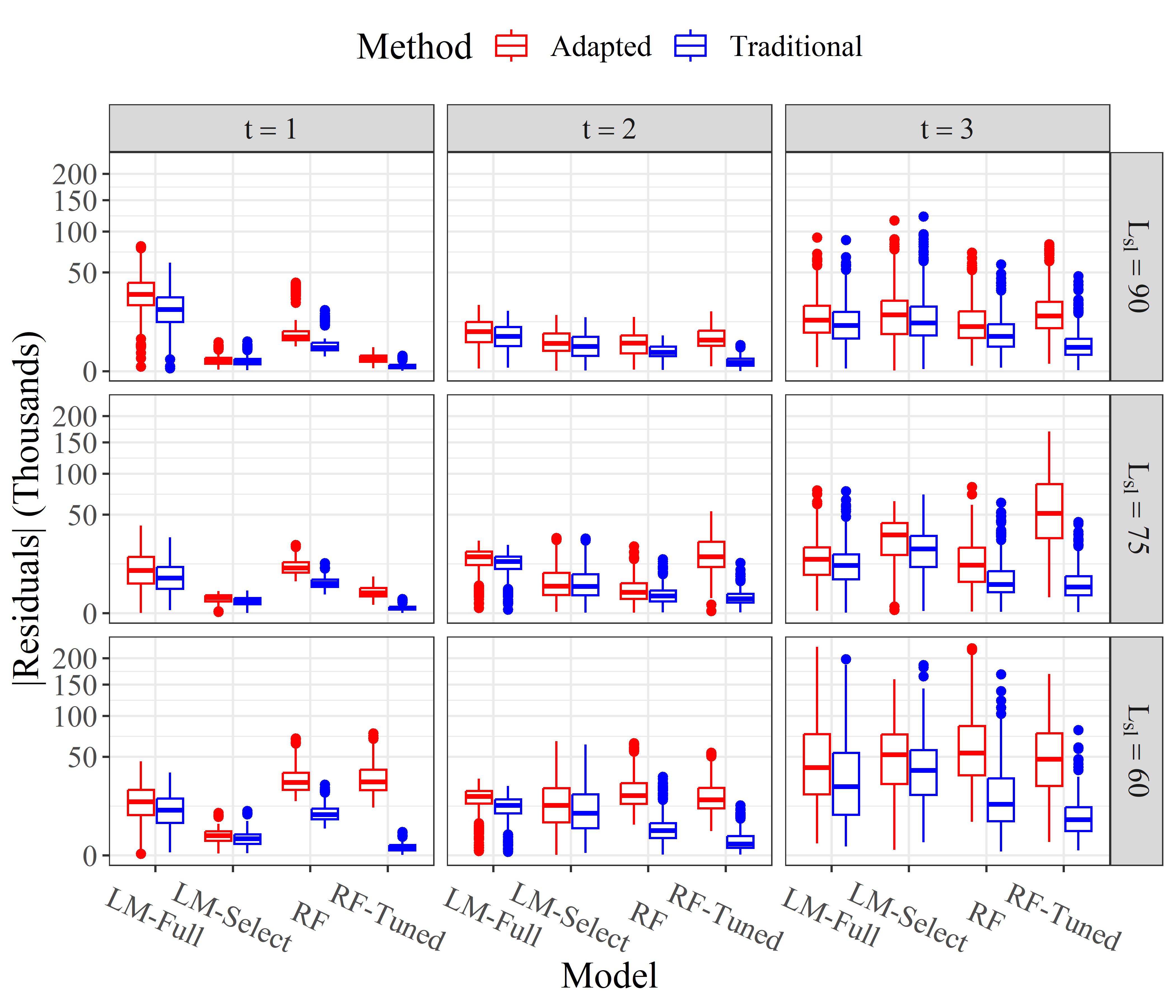}
    \caption{Boxplot of absolute errors from the traditional and adapted cross-validation as separated by $t$ and $L_{cl}$ group combination.}
    \label{fig:aeplot}
\end{figure}

These summary statistics are explored in more detail in Figure \ref{fig:aeplot}. This figure illustrates the regions of the $t$ and $L_{cl}$ parameter space prone to inaccuracies. Of note is the similarity of the traditional and adapted boxplots for the linear regression models, in contrast to the markedly different boxplots for the RF models. This shows the inability of the RF models to appropriately generalize predictions to certain combinations of the $t$ and $L_{cl}$ parameter space, even for observations within the center of the parameter space where appropriate extrapolation may have been expected.    

The key insight from Table \ref{tab:rmse} is that the LMs behave roughly similarly in both the traditional and adapted cross-validation scenarios, while the RF model performance degrades sharply when moving to the adapted cross-validation scenario. In fact, the LMs actually outperform the RF models in the adapted cross-validation scenario. Similar results are observed across other top-performing machine learning models included in the stressor output. 

This example reiterates the dangers of model over-fitting. However, the example goes beyond over-fitting and illustrates that commonly used measures of accuracy may provide inappropriate measures of model performance, regardless of model over-fitting. As such, it is important for researchers to evaluate model accuracy across a variety of scenarios, rather than trying to summarize accuracy as a single quantity.

\subsection{Example 2: Model interpretation and engineering intuition}
The RC wall database was used to examine two modeling scenarios: (1) \textit{Scenario 1}: The ML models neglect a physically important feature. This scenario corresponds to the "variable omission" case, when the developer compensates for the missed feature by introducing other parameters in model development, (2) \textit{Scenario 2}: The ML models obtain similar accuracy with different subsets of features (i.e., underspecification). This scenario mimics when the developers examine additional features (besides those that domain knowledge suggests) to improve model accuracy and discover new features that are important for the predictive model.  
An illustration of this point is made using the two ML models: random forest (RF) and support vector machines (SVM). These models are trained using a standard ML pipeline The database is split into 70\% for the train set and 30\% for the test set. The RF model is tuned only using three hyperparameters: the number of trees, the maximum depth of individual trees, and the minimum number of samples in each split (i.e., leaf) of a tree. In contrast, the regularization term and width of the kernel are selected as hyperparameters of SVM with radial basis function kernel.  The hyperparameter values are obtained through a combination of random and grid searches in hyperparameter space using a 3-fold cross-validation. It should be noted that to serve the illustrative nature of this example, the number of examined hyperparameters and data transformation is kept small. Model accuracy is measured using $R^2$ and root mean squared error (RMSE). 
\subsubsection{Variable omission bias}
For the first scenario, one physically important feature is excluded in each ML model. Such a scenario happens when an ML model is developed based on a limited database, or when an appropriately developed ML will be deployed for a case where real-world data is not available for a given feature of model. Previous prescriptive equations by Abdullah and Wallace \cite{abdullah2019} indicate that the wall drift is related to $\lambda_b$ and $\dfrac{\nu_{max}}{\sqrt{f'_c}}$ as follows:

\begin{equation}
      \frac{\delta_c}{h_w} = 3.85-\frac{\lambda_b}{\alpha} - \frac{\nu_{max}}{10\sqrt{f'_c}}
\end{equation}

\begin{figure}[t]
   \centering
        \includegraphics[width=15 cm]{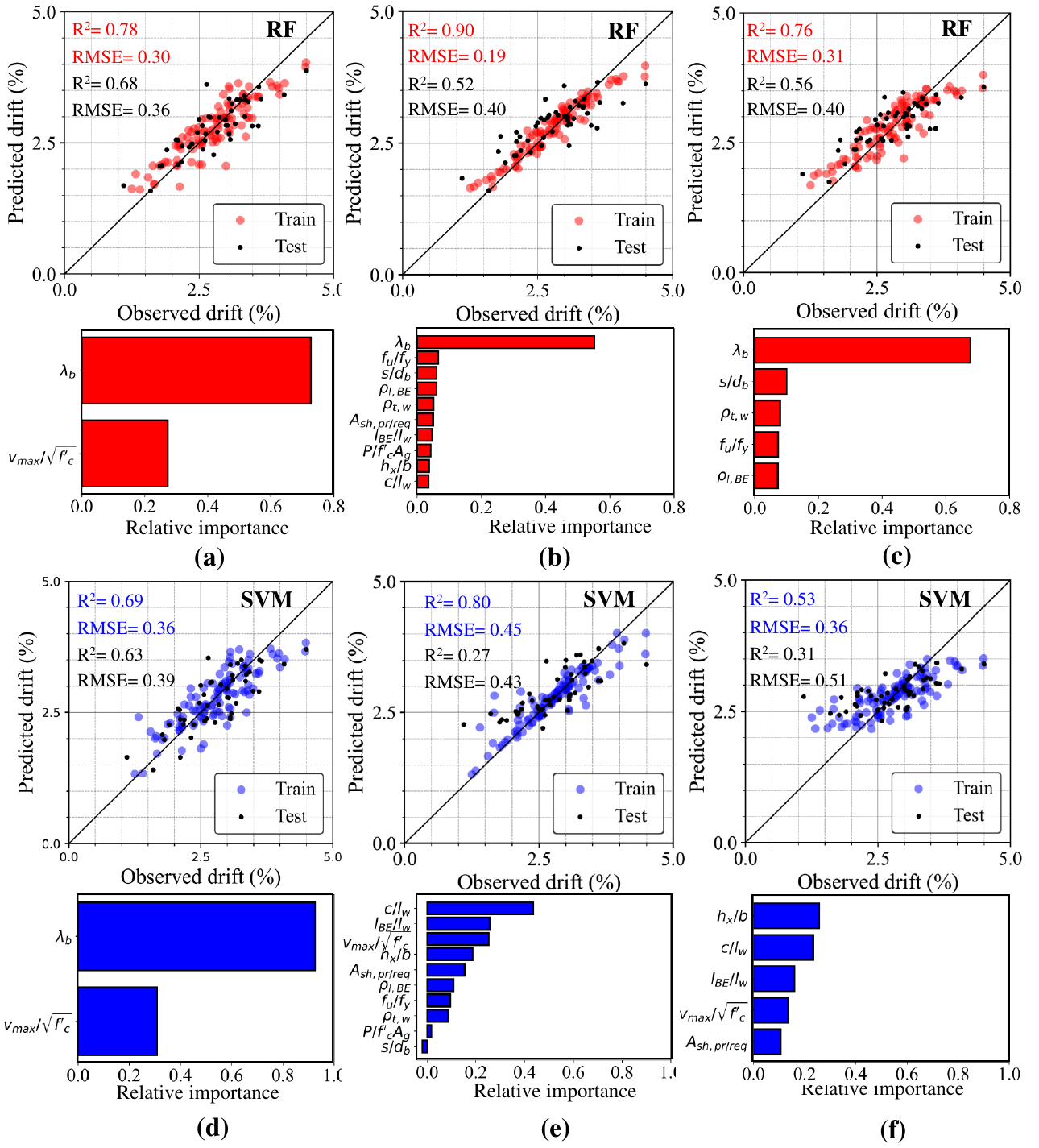}
    \caption {Effect of variable omission: (a), (b), and (c) compare RF models with only the physically important features, all features except for one physically important one, and one that is built based on best features of a model without a physically important feature, respectively. (d), (e) and (f) compare similarly developed SVM models. }
    \label{VarOmit}
\end{figure}

Figure \ref{VarOmit}.a shows an RF model that includes the physically important features of $\lambda_b$ and $\dfrac{\nu_{max}}{\sqrt{f'_c}}$ as input. The model has an $R^2$ and RMSE of 0.78 and 0.3\% on the train set, whereas the same metrics change to 0.68 and 0.36\% for the test set. Figure \ref{VarOmit}.b shows the scenario when the developer uses a feature set comprising all other parameters except for the physically important feature of $\dfrac{\nu_{max}}{\sqrt{f'_c}}$. It can be observed this model now exhibits overfitting, as shown by substantially better performance on the train set while showing lower accuracy metrics on the test set (i.e., 23.5\% reduction of $R^2$ and 11.1\% increase of RMSE compared to \ref{VarOmit}.a). As discussed in Section \ref{Generalizability}, to overcome the overfitting, the developer might reduce the model complexity by only using fewer features, such as the most important features from model \ref{VarOmit}.b (i.e., \{$\lambda_b,s/d_b,\rho_{l,BE},\rho_{t,w},f_u/f_y$\}). As Figure \ref{VarOmit}.c shows, this model would reduce the difference in accuracy metrics between the train and test set. However, the performance on the test set is not improved (as gauged by the same RMSE as Figure \ref{VarOmit}.b) or has negligible change (7.7\% improvement over $R^2$). In summary, an RF model that ignores a physically important feature can be expected to perform poorly in deployment, and the introduction of additional features in the development stage cannot compensate for the omitted variables. 

Figures \ref{VarOmit}.d-\ref{VarOmit}.f shows a similar trend for SVM. The exclusion of $\lambda_b$ led to overfitting, where the model with all features except for $\lambda_b$ achieved 57.1\% lower $R^2$ and 10.2\% higher RMSE on test set, respectively. Here, developing an SVM model using the most important features from Figure \ref{VarOmit}.e led to poorer performance, as shown by Figure \ref{VarOmit}.f. This observation supports the argument that using feature importance algorithms (such as permutation-based feature importance) cannot account for the effect of omitted variables. The lower $R^2$ of the model on the train set (as indicated by the horizontal spread of predictions) indicates that by removing all seemingly unimportant features, their interactions were removed, and model's ability to learn the underlying relationship was decreased. 

\subsubsection{Underspecification}
The second scenario highlights underspecification in ML development. For this scenario, a physically meaningful feature ($\lambda_b$ or ($\nu_{max}/\sqrt{f'_c}$)) is kept in both models, and other features are introduced to improve model accuracy. This is a common scenario in ML development, where the modeler knows several important features from the literature and aims to create a more accurate ML model by examining other candidate features. Shapley additive values (a model-agnostic interpretability technique) were used to explain feature importance and their relationship to the response.

\begin{figure}[t]
   \centering
        \includegraphics[width=14 cm]{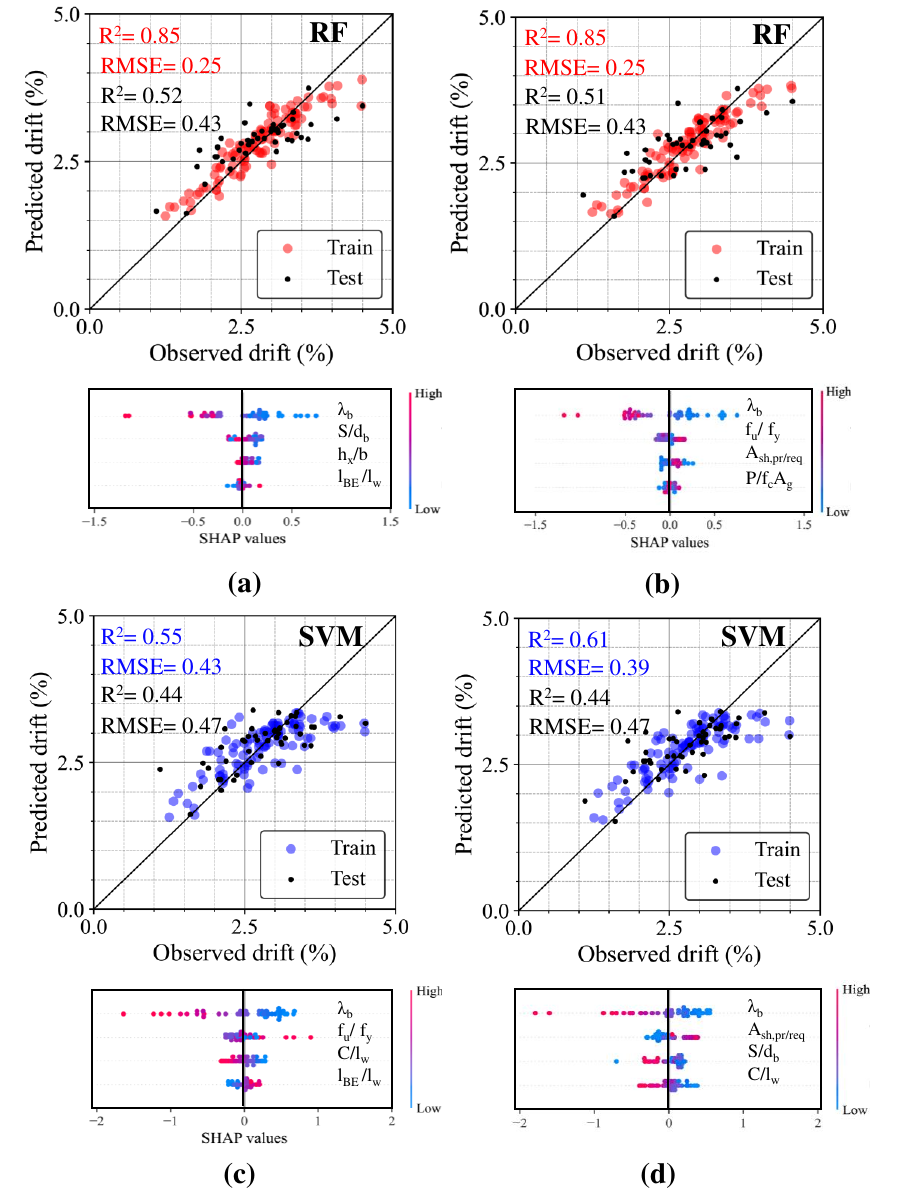}
    \caption {Underspecification in ML models: (a) and (b) show RF model with similar performance on test set using different  input features , whereas (c) and (d) shows SVM model with different features achieving similar accuracy on test}
    \label{underspecification}
\end{figure}

Figures \ref{underspecification}.a and \ref{underspecification}.b show two different RF models (with different input features) that achieved similar accuracy on both train and test sets. The $R^2$ of both models are 0.85 on the train set, whereas, for the test set, these values change to 0.51 and 0.52. Similarly, model RMSE is 0.25 and 0.43 on the train and test set for both models. In contrast, the model showed in \ref{underspecification}.a uses $\{\lambda_b, s/d_b, h_x/b, l_{BE}/l_w\}$, whereas model showed in \ref{underspecification}.b uses $\{\lambda_b, f_u/f_y, A_{sh,provided}/A_{sh,required}, P/f'_cA_g \}$. Shapely plots indicate that while $\lambda_b$ is the most important feature, $s/d_b$, $h_x/b$, and $l_{BE}/l_w$ have similar importance for the first model, whereas the $f_u/f_y$ and $A_{sh,provided}/A_{sh,required}$ are equally important for prediction in the second model, showing underspecification in ML development. Similarly, \ref{underspecification}.c and \ref{underspecification}.d shows two SVM models with different features that achieve similar accuracy on the test set. 

In addition, it is possible that Shapely plots show inconsistency between response and feature relationships in different ML algorithms. For example, while shapely plots show similar decreasing and increasing relationship between drift and $\lambda_b$, $f_u/f_y$ and $C/l_w$ over the SVM and RF models, the relationship between $A_{sh,provided}/A_{sh,required}$ and drift is captured differently between the RF and SVM models. As shown in \ref{underspecification}.b and \ref{underspecification}.d, an increase in the ratio of provided to the required area of the boundary transverse reinforcement increases drift in SVM model, whereas it results in a decrease in drift for larger ratio values for the RF model.

The presented example exhibits the limitations of relying on purely data-driven approaches to gain insights from ML models, particularly when a physically important feature is absent, or when underspecification limits the explanation (and generalizability) of a given model. In this example, it is most likely that none of the features except for $\lambda_b$ and $\dfrac{\nu_{max}}{\sqrt{f'_c}}$ can truly explain wall drift. As such, while their inclusion can lead to models with higher accuracy on the training set, such overfit models (with additional features) will perform poorly on a new dataset. Performing feature importance will provide misleading insights on their importance, and even using such insights to trim models will not lead to better models, highlighting the importance of being well-informed on known physical relationships for proper ML development

\section{Conclusion}

This study investigated the possible pitfalls for developing ML models that fail at deployment due to generalizability and explainability challenges. A wide range of issues ranging from overfitting, underrepresentiveness of training data, and variable omission bias to ML model underspecification were discussed. In addition, illustrative examples were presented using one finite element simulations dataset (shear strength of CFS channels), and one experimental study dataset (drift capacity of RC shear walls). Using these examples, the aforementioned issues were mapped into the structural engineering domain, and results were discussed from a statistical perspective. The following summarizes insights from the presented discussion for developing ML-based solutions that are more suitable for deployment in the structural engineering field: 

\begin{enumerate}
     \item Multiple methods for assessing model accuracy should be implemented. As shown in the CFS channel illustration, unlike the traditional cross-validation approach, adaptive cross-validation (i.e., performing cross-validation on regions demarcated by significant features) was able to highlight the generalizability issues of developed RF methods.      
    \item Models with varying complexity should be examined as sometimes ML models with lower complexity could have a higher value for deployment due to better generalizability or better explainability. For the CFS channel database, linear regression models outperformed random forest models by providing consistent prediction irrespective of the method of cross-validation.
    \item ML models' generalizability and explainability rely on the inclusion of physically-important features (and relationships) and ensuring that the ML model reflects that information. Overt inclusion of marginally significant features cannot compensate for variable omission bias, and led to complex, yet fragile, statistical models with poor deployment values. In the RC shear wall database, models including all available features except for one physically important variable led to relatively poor performance on new data.   
    \item Common feature importance methods are sensitive to correlation between features, cannot account for the effect of variable omission, and are not an indicator of causality. Using such methods for underspecified models can lead to misleading insights on the underlying relationship between response and features.  
    \item Contrary to the common fallacy of "more data is better" for ML development, additional data will not lead to deployable ML models if that data is not an appropriate representation of real-world data distribution with respect to selected features.
\end{enumerate}

\subsection{Data Availability Statement}

Some or all data, models, or code that support the findings of this study are available from the corresponding author upon reasonable request (list items).
\subsection{Acknowledgments}

%
%
\bibliography{ascexmpl-new.bbl}

\begin{thebibliography}{}

\bibitem[\protect\citeauthoryear{}{Abdullah and Wallace}{2019}]{abdullah2019}
Abdullah, S.~A. and Wallace, J.~W. (2019).
\newblock ``Drift capacity of reinforced concrete structural walls with special
  boundary elements.''\ {\em ACI Structural Journal}, 116(1), 183.

\bibitem[\protect\citeauthoryear{}{Aladsani et~al.\@}{2022}]{aladsani2022}
Aladsani, M.~A., Burton, H., Abdullah, S.~A., and Wallace, J.~W. (2022).
\newblock ``Explainable machine learning model for predicting drift capacity of
  reinforced concrete walls..''\ {\em ACI Structural Journal}, 119(3).

\bibitem[\protect\citeauthoryear{}{Ali}{2020}]{PyCaret}
Ali, M. (2020).
\newblock {\em PyCaret: An open source, low-code machine learning library in
  Python}, $<$https://www.pycaret.org$>$\ (April).
\newblock PyCaret version 1.0.0. Accessed May 17, 2023.

\bibitem[\protect\citeauthoryear{}{Allen et~al.\@}{2023}]{allen2023}
Allen, G.~I., Gan, L., and Zheng, L. (2023).
\newblock ``Interpretable machine learning for discovery: Statistical
  challenges$\backslash$\& opportunities.''\ {\em arXiv preprint
  arXiv:2308.01475}.

\bibitem[\protect\citeauthoryear{}{Baier et~al.\@}{2019}]{baier2019}
Baier, L., J{\"o}hren, F., and Seebacher, S. (2019).
\newblock ``Challenges in the deployment and operation of machine learning in
  practice..''\ {\em ECIS}, Vol.~1.

\bibitem[\protect\citeauthoryear{}{Bashar and Torres-Machi}{2022}]{bashar2022}
Bashar, M.~Z. and Torres-Machi, C. (2022).
\newblock ``Deep learning for estimating pavement roughness using synthetic
  aperture radar data.''\ {\em Automation in Construction}, 142, 104504.

\bibitem[\protect\citeauthoryear{}{Burton}{2023}]{burton2023causal}
Burton, H. (2023).
\newblock ``Causal inference on observational data: Opportunities and
  challenges in earthquake engineering.''\ {\em Earthquake Spectra}, 39(1),
  54--76.

\bibitem[\protect\citeauthoryear{}{Burton and Baker}{2023}]{burton2023}
Burton, H.~V. and Baker, J.~W. (2023).
\newblock ``Evaluating the effectiveness of ground motion intensity measures
  through the lens of causal inference.''\ {\em Earthquake Engineering \&
  Structural Dynamics}.

\bibitem[\protect\citeauthoryear{}{Cakiroglu et~al.\@}{2022}]{cakiroglu2022}
Cakiroglu, C., Islam, K., Bekda{\c{s}}, G., Isikdag, U., and Mangalathu, S.
  (2022).
\newblock ``Explainable machine learning models for predicting the axial
  compression capacity of concrete filled steel tubular columns.''\ {\em
  Construction and Building Materials}, 356, 129227.

\bibitem[\protect\citeauthoryear{}{Chernozhukov
  et~al.\@}{2021}]{chernozhukov2021}
Chernozhukov, V., Cinelli, C., Newey, W.~K., Sharma, A., and Syrgkanis, V.
  (2021).
\newblock ``Omitted variable bias in machine learned causal models.''\ {\em
  Report no.}, cemmap working paper.

\bibitem[\protect\citeauthoryear{}{Clarke}{2005}]{clarke2005}
Clarke, K.~A. (2005).
\newblock ``The phantom menace: Omitted variable bias in econometric
  research.''\ {\em Conflict management and peace science}, 22(4), 341--352.

\bibitem[\protect\citeauthoryear{}{D'Amour et~al.\@}{2020}]{d2020}
D'Amour, A.~N., Heller, K., Moldovan, D., Adlam, B., Alipanahi, B., Beutel, A.,
  Chen, C., Deaton, J., Eisenstein, J., Hoffman, M.~D., et~al.\@ (2020).
\newblock ``Underspecification presents challenges for credibility in modern
  machine learning.

\bibitem[\protect\citeauthoryear{}{Debeer and Strobl}{2020}]{debeer2020}
Debeer, D. and Strobl, C. (2020).
\newblock ``Conditional permutation importance revisited.''\ {\em BMC
  bioinformatics}, 21(1), 1--30.

\bibitem[\protect\citeauthoryear{}{Degtyarev}{2021}]{degtyarev2021}
Degtyarev, V.~V. (2021).
\newblock ``Neural networks for predicting shear strength of cfs channels with
  slotted webs.''\ {\em Journal of Constructional Steel Research}, 177, 106443.

\bibitem[\protect\citeauthoryear{}{Eche et~al.\@}{2021}]{eche2021}
Eche, T., Schwartz, L.~H., Mokrane, F.-Z., and Dercle, L. (2021).
\newblock ``Toward generalizability in the deployment of artificial
  intelligence in radiology: role of computation stress testing to overcome
  underspecification.''\ {\em Radiology: Artificial Intelligence}, 3(6),
  e210097.

\bibitem[\protect\citeauthoryear{}{Esteghamati and
  Flint}{2021}]{esteghamati2021}
Esteghamati, M.~Z. and Flint, M.~M. (2021).
\newblock ``Developing data-driven surrogate models for holistic
  performance-based assessment of mid-rise rc frame buildings at early
  design.''\ {\em Engineering Structures}, 245, 112971.

\bibitem[\protect\citeauthoryear{}{Esteghamati and
  Flint}{2023}]{esteghamati2023all}
Esteghamati, M.~Z. and Flint, M.~M. (2023).
\newblock ``Do all roads lead to rome? a comparison of knowledge-based,
  data-driven, and physics-based surrogate models for performance-based early
  design.''\ {\em Engineering Structures}, 286, 116098.

\bibitem[\protect\citeauthoryear{}{Esteghamati
  et~al.\@}{2023}]{esteghamati2023}
Esteghamati, M.~Z., Gernay, T., and Banerji, S. (2023).
\newblock ``Evaluating fire resistance of timber columns using explainable
  machine learning models.''\ {\em Engineering Structures}, 296, 116910.

\bibitem[\protect\citeauthoryear{}{Feng et~al.\@}{2021}]{feng2021}
Feng, D.-C., Wang, W.-J., Mangalathu, S., and Taciroglu, E. (2021).
\newblock ``Interpretable xgboost-shap machine-learning model for shear
  strength prediction of squat rc walls.''\ {\em Journal of Structural
  Engineering}, 147(11), 04021173.

\bibitem[\protect\citeauthoryear{}{Fisher et~al.\@}{2019}]{fisher2019}
Fisher, A., Rudin, C., and Dominici, F. (2019).
\newblock ``All models are wrong, but many are useful: Learning a variable's
  importance by studying an entire class of prediction models
  simultaneously..''\ {\em J. Mach. Learn. Res.}, 20(177), 1--81.

\bibitem[\protect\citeauthoryear{}{Ge et~al.\@}{2014}]{ge2014}
Ge, L., Gao, J., Ngo, H., Li, K., and Zhang, A. (2014).
\newblock ``On handling negative transfer and imbalanced distributions in
  multiple source transfer learning.''\ {\em Statistical Analysis and Data
  Mining: The ASA Data Science Journal}, 7(4), 254--271.

\bibitem[\protect\citeauthoryear{}{Haycock and Bean}{2023}]{stressor}
Haycock, S. and Bean, B. (2023).
\newblock {\em stressor: Algorithms for testing models under stress},
  $<$https://github.com/beanb2/stressor$>$.
\newblock R package version 0.1.0.

\bibitem[\protect\citeauthoryear{}{Hooker et~al.\@}{2021}]{hooker2021}
Hooker, G., Mentch, L., and Zhou, S. (2021).
\newblock ``Unrestricted permutation forces extrapolation: variable importance
  requires at least one more model, or there is no free variable importance.''\
  {\em Statistics and Computing}, 31, 1--16.

\bibitem[\protect\citeauthoryear{}{Hwang et~al.\@}{2021}]{hwang2021}
Hwang, S.-H., Mangalathu, S., Shin, J., and Jeon, J.-S. (2021).
\newblock ``Machine learning-based approaches for seismic demand and collapse
  of ductile reinforced concrete building frames.''\ {\em Journal of Building
  Engineering}, 34, 101905.

\bibitem[\protect\citeauthoryear{}{Issa et~al.\@}{2023}]{issa2023}
Issa, O., Silva-Lopez, R., Baker, J.~W., and Burton, H.~V. (2023).
\newblock ``Machine-learning-based optimization framework to support
  recovery-based design.''\ {\em Earthquake Engineering \& Structural
  Dynamics}.

\bibitem[\protect\citeauthoryear{}{Koh and Blum}{2022}]{koh2022}
Koh, H. and Blum, H.~B. (2022).
\newblock ``Machine learning-based sensitivity of steel frames with highly
  imbalanced and high-dimensional data.''\ {\em Engineering Structures}, 259,
  114126.

\bibitem[\protect\citeauthoryear{}{Kourehpaz and
  Molina~Hutt}{2022}]{kourehpaz2022}
Kourehpaz, P. and Molina~Hutt, C. (2022).
\newblock ``Machine learning for enhanced regional seismic risk assessments.''\
  {\em Journal of Structural Engineering}, 148(9), 04022126.

\bibitem[\protect\citeauthoryear{}{Li et~al.\@}{2022}]{li2022}
Li, Z., Yoon, J., Zhang, R., Rajabipour, F., Srubar~III, W.~V., Dabo, I., and
  Radli{\'n}ska, A. (2022).
\newblock ``Machine learning in concrete science: applications, challenges, and
  best practices.''\ {\em npj Computational Materials}, 8(1), 127.

\bibitem[\protect\citeauthoryear{}{Lundberg and Lee}{2017}]{lundberg2017}
Lundberg, S.~M. and Lee, S.-I. (2017).
\newblock ``A unified approach to interpreting model predictions.''\ {\em
  Advances in neural information processing systems}, 30.

\bibitem[\protect\citeauthoryear{}{Luo and Paal}{2021}]{luo2021}
Luo, H. and Paal, S.~G. (2021).
\newblock ``Reducing the effect of sample bias for small data sets with
  double-weighted support vector transfer regression.''\ {\em Computer-Aided
  Civil and Infrastructure Engineering}, 36(3), 248--263.

\bibitem[\protect\citeauthoryear{}{Luo and Paal}{2023}]{luo2023}
Luo, H. and Paal, S.~G. (2023).
\newblock ``A novel outlier-insensitive local support vector machine for robust
  data-driven forecasting in engineering.''\ {\em Engineering with Computers},
  1--19.

\bibitem[\protect\citeauthoryear{}{Mathrani et~al.\@}{2021}]{mathrani2021}
Mathrani, A., Susnjak, T., Ramaswami, G., and Barczak, A. (2021).
\newblock ``Perspectives on the challenges of generalizability, transparency
  and ethics in predictive learning analytics.''\ {\em Computers and Education
  Open}, 2, 100060.

\bibitem[\protect\citeauthoryear{}{Mohammadi et~al.\@}{2023}]{mohammadi2023}
Mohammadi, P., Rashidi, A., Malekzadeh, M., and Tiwari, S. (2023).
\newblock ``Evaluating various machine learning algorithms for automated
  inspection of culverts.''\ {\em Engineering Analysis with Boundary Elements},
  148, 366--375.

\bibitem[\protect\citeauthoryear{}{Molnar et~al.\@}{2020}]{molnar2020}
Molnar, C., Casalicchio, G., and Bischl, B. (2020).
\newblock ``Interpretable machine learning--a brief history, state-of-the-art
  and challenges.''\ {\em Joint European conference on machine learning and
  knowledge discovery in databases}, Springer,  417--431.

\bibitem[\protect\citeauthoryear{}{Molnar et~al.\@}{2023}]{molnar2023}
Molnar, C., K{\"o}nig, G., Bischl, B., and Casalicchio, G. (2023).
\newblock ``Model-agnostic feature importance and effects with dependent
  features: a conditional subgroup approach.''\ {\em Data Mining and Knowledge
  Discovery},  1--39.

\bibitem[\protect\citeauthoryear{}{Naser}{2022}]{naser2022}
Naser, M. (2022).
\newblock ``Causality in structural engineering: discovering new knowledge by
  tying induction and deduction via mapping functions and explainable
  artificial intelligence.''\ {\em AI in Civil Engineering}, 1(1), 6.

\bibitem[\protect\citeauthoryear{}{Naser and
  {\c{C}}ift{\c{c}}io{\u{g}}lu}{2023}]{naser2023}
Naser, M. and {\c{C}}ift{\c{c}}io{\u{g}}lu, A.~{\"O}. (2023).
\newblock ``Causal discovery and inference for evaluating fire resistance of
  structural members through causal learning and domain knowledge.''\ {\em
  Structural Concrete}.

\bibitem[\protect\citeauthoryear{}{Naser and Kodur}{2022}]{naser2022e}
Naser, M. and Kodur, V. (2022).
\newblock ``Explainable machine learning using real, synthetic and augmented
  fire tests to predict fire resistance and spalling of rc columns.''\ {\em
  Engineering Structures}, 253, 113824.

\bibitem[\protect\citeauthoryear{}{Pan and Yang}{2009}]{pan2009}
Pan, S.~J. and Yang, Q. (2009).
\newblock ``A survey on transfer learning.''\ {\em IEEE Transactions on
  knowledge and data engineering}, 22(10), 1345--1359.

\bibitem[\protect\citeauthoryear{}{{Python Core Team}}{2019}]{python38}
{Python Core Team} (2019).
\newblock {\em {Python: A dynamic, open source programming language}}.
\newblock {Python Software Foundation}, $<$https://www.python.org/$>$.
\newblock Python version 3.8.

\bibitem[\protect\citeauthoryear{}{Qin and Naser}{2023}]{qin2023}
Qin, Z. and Naser, M. (2023).
\newblock ``Machine learning and model driven bayesian uncertainty
  quantification in suspended nonstructural systems.''\ {\em Reliability
  Engineering \& System Safety}, 237, 109392.

\bibitem[\protect\citeauthoryear{}{{R Core Team}}{2023}]{R2023}
{R Core Team} (2023).
\newblock {\em R: A Language and Environment for Statistical Computing}.
\newblock R Foundation for Statistical Computing, Vienna, Austria,
  $<$https://www.R-project.org/$>$.

\bibitem[\protect\citeauthoryear{}{Ramspek et~al.\@}{2021}]{ramspek2021}
Ramspek, C.~L., Jager, K.~J., Dekker, F.~W., Zoccali, C., and van Diepen, M.
  (2021).
\newblock ``External validation of prognostic models: what, why, how, when and
  where?.''\ {\em Clinical Kidney Journal}, 14(1), 49--58.

\bibitem[\protect\citeauthoryear{}{Rudin et~al.\@}{2022}]{rudin2022}
Rudin, C., Chen, C., Chen, Z., Huang, H., Semenova, L., and Zhong, C. (2022).
\newblock ``Interpretable machine learning: Fundamental principles and 10 grand
  challenges.''\ {\em Statistic Surveys}, 16, 1--85.

\bibitem[\protect\citeauthoryear{}{Sculley et~al.\@}{2015}]{sculley2015}
Sculley, D., Holt, G., Golovin, D., Davydov, E., Phillips, T., Ebner, D.,
  Chaudhary, V., Young, M., Crespo, J.-F., and Dennison, D. (2015).
\newblock ``Hidden technical debt in machine learning systems.''\ {\em Advances
  in neural information processing systems}, 28.

\bibitem[\protect\citeauthoryear{}{Shapley et~al.\@}{1953}]{shapley1953}
Shapley, L.~S. et~al.\@ (1953).
\newblock ``A value for n-person games.

\bibitem[\protect\citeauthoryear{}{Soleimani-Babakamali
  et~al.\@}{2023}]{soleimani2023}
Soleimani-Babakamali, M.~H., Soleimani-Babakamali, R., Nasrollahzadeh, K.,
  Avci, O., Kiranyaz, S., and Taciroglu, E. (2023).
\newblock ``Zero-shot transfer learning for structural health monitoring using
  generative adversarial networks and spectral mapping.''\ {\em Mechanical
  Systems and Signal Processing}, 198, 110404.

\bibitem[\protect\citeauthoryear{}{Strobl et~al.\@}{2008}]{strobl2008}
Strobl, C., Boulesteix, A.-L., Kneib, T., Augustin, T., and Zeileis, A. (2008).
\newblock ``Conditional variable importance for random forests.''\ {\em BMC
  bioinformatics}, 9, 1--11.

\bibitem[\protect\citeauthoryear{}{Vabalas et~al.\@}{2019}]{vabalas2019}
Vabalas, A., Gowen, E., Poliakoff, E., and Casson, A.~J. (2019).
\newblock ``Machine learning algorithm validation with a limited sample
  size.''\ {\em PloS one}, 14(11), e0224365.

\bibitem[\protect\citeauthoryear{}{Varma and Simon}{2006}]{varma2006}
Varma, S. and Simon, R. (2006).
\newblock ``Bias in error estimation when using cross-validation for model
  selection.''\ {\em BMC bioinformatics}, 7(1), 1--8.

\bibitem[\protect\citeauthoryear{}{Wakjira et~al.\@}{2022}]{wakjira2022}
Wakjira, T.~G., Ibrahim, M., Ebead, U., and Alam, M.~S. (2022).
\newblock ``Explainable machine learning model and reliability analysis for
  flexural capacity prediction of rc beams strengthened in flexure with
  frcm.''\ {\em Engineering Structures}, 255, 113903.

\bibitem[\protect\citeauthoryear{}{Watson}{2022}]{watson2022}
Watson, D.~S. (2022).
\newblock ``Conceptual challenges for interpretable machine learning.''\ {\em
  Synthese}, 200(2), 65.

\bibitem[\protect\citeauthoryear{}{Xu et~al.\@}{2022}]{xu2022}
Xu, J.-G., Feng, D.-C., Mangalathu, S., and Jeon, J.-S. (2022).
\newblock ``Data-driven rapid damage evaluation for life-cycle seismic
  assessment of regional reinforced concrete bridges.''\ {\em Earthquake
  Engineering \& Structural Dynamics}, 51(11), 2730--2751.

\bibitem[\protect\citeauthoryear{}{Youssef et~al.\@}{2023}]{youssef2023}
Youssef, A., Pencina, M., Thakur, A., Zhu, T., Clifton, D., and Shah, N.~H.
  (2023).
\newblock ``All models are local: time to replace external validation with
  recurrent local validation.''\ {\em arXiv preprint arXiv:2305.03219}.

\bibitem[\protect\citeauthoryear{}{Yu et~al.\@}{2022}]{yu2022}
Yu, A.~C., Mohajer, B., and Eng, J. (2022).
\newblock ``External validation of deep learning algorithms for radiologic
  diagnosis: a systematic review.''\ {\em Radiology: Artificial Intelligence},
  4(3), e210064.

\bibitem[\protect\citeauthoryear{}{Yuan et~al.\@}{2023}]{yuan2023}
Yuan, X., Smith, A., Moreu, F., Sarlo, R., Lippitt, C.~D., Hojati, M.,
  Alampalli, S., and Zhang, S. (2023).
\newblock ``Automatic evaluation of rebar spacing and quality using lidar data:
  Field application for bridge structural assessment.''\ {\em Automation in
  Construction}, 146, 104708.

\bibitem[\protect\citeauthoryear{}{Zhao et~al.\@}{2023}]{zhao2023}
Zhao, P., Liao, W., Huang, Y., and Lu, X. (2023).
\newblock ``Intelligent design of shear wall layout based on graph neural
  networks.''\ {\em Advanced Engineering Informatics}, 55, 101886.

\end{thebibliography}

\end{document}